\PassOptionsToPackage{prologue,dvipsnames,table}{xcolor}
\documentclass[acmtog, screen, nonacm]{acmart}
\usepackage{tikz}
\usetikzlibrary{shapes, arrows.meta, decorations.markings}
\usepackage{float}
\usepackage{caption}
\usepackage{booktabs} 
\usepackage{bbm}
\usepackage{hyperref}
\usepackage{enumitem}
\usepackage{booktabs}
\usepackage{siunitx}
\usepackage{nicefrac}
\usepackage{algorithm}
\usepackage{algpseudocode}
\usepackage[normalem]{ulem}
\usepackage{empheq} 
\usepackage{algorithm}
\usepackage{algpseudocode}
\usepackage{amsmath}
\usepackage{acronym}

\definecolor{codegreen}{rgb}{0,0.6,0}
\definecolor{codegray}{rgb}{0.5,0.5,0.5}
\definecolor{codepurple}{rgb}{0.58,0,0.82}
\definecolor{backcolour}{rgb}{0.96,0.96,0.98}
\usepackage{listings}
\usepackage{xcolor}
\setcitestyle{square}
\setcopyright{none}

\makeatletter
\let\@authorsaddresses\@empty
\makeatother

\title{GarmentDreamer: 3DGS Guided Garment Synthesis with Diverse Geometry and Texture Details}

\author{Boqian Li$^{1,2*}$}
\author{Xuan Li$^{1*}$}
\author{Ying Jiang$^{1, 3*}$}
\author{Tianyi Xie$^{1}$}
\author{Feng Gao$^{4}$}
\author{Huamin Wang$^{5}$}
\author{Yin Yang$^{2}$}
\author{Chenfanfu Jiang$^{1}$}

\usepackage{algorithm}
\usepackage{amsmath}

\usepackage{algorithmicx}
\usepackage{algpseudocode}
\usepackage{multirow}
\usepackage{caption}
\usepackage{subcaption,bm}
\usepackage{stmaryrd}
\usepackage{wrapfig}
\usepackage{{threeparttable}}
\algdef{SE}[DOWHILE]{Do}{doWhile}{\algorithmicdo}[1]{\algorithmicwhile\ #1}
\AtBeginDocument{%
  \providecommand\BibTeX{{%
    \normalfont B\kern-0.5em{\scshape i\kern-0.25em b}\kern-0.8em\TeX}}}

\begin{document}

\acrodef{sds}[SDS]{Score Distillation Sampling}
\acrodef{vsd}[VSD]{Variational Score Distillation}
\acrodef{udf}[UDF]{Unsigned Distance Field}
\acrodef{cv}[CV]{Computer Vision}
\acrodef{cg}[CG]{Computer Graphics}
\acrodef{aigc}[AIGC]{Artificial Intelligence Generated Content}
\acrodef{vr}[VR]{Virtual Reality}
\acrodef{mr}[MR]{Mixed Reality}
\acrodef{nerf}[NeRF]{Neural Radiance Fields}
\acrodef{gs}[GS]{Gaussian Splatting}
\acrodef{vae}[VAE]{Variational Encoder}
\newcommand\blfootnote[1]{
    \begingroup
    \renewcommand\thefootnote{}\footnote{#1}
    \addtocounter{footnote}{-1}
    \endgroup
}

\begin{abstract}

\blfootnote{* indicates equal contributions. \\Affiliations: $^1$UCLA, $^2$Utah, $^3$HKU, $^4$Amazon (This work is not related to F. Gao’s position at Amazon.), $^5$Style3D Research}

Traditional 3D garment creation is labor-intensive, involving sketching, modeling, UV mapping, and texturing, which are time-consuming and costly. Recent advances in diffusion-based generative models have enabled new possibilities for 3D garment generation from text prompts, images, and videos. However, existing methods either suffer from inconsistencies among multi-view images or require additional processes to separate cloth from the underlying human model. In this paper, we propose GarmentDreamer, a novel method that leverages 3D Gaussian Splatting (GS) as guidance to generate wearable, simulation-ready 3D garment meshes from text prompts. In contrast to using multi-view images directly predicted by generative models as guidance, our 3DGS guidance ensures consistent optimization in both garment deformation and texture synthesis. Our method introduces a novel garment augmentation module, guided by normal and RGBA information, and employs implicit Neural Texture Fields (NeTF) combined with Variational Score Distillation (VSD) to generate diverse geometric and texture details. We validate the effectiveness of our approach through comprehensive qualitative and quantitative experiments, showcasing the superior performance of GarmentDreamer over state-of-the-art alternatives. Our project page is available at \url{https://xuan-li.github.io/GarmentDreamerDemo/}.
\end{abstract}


\begin{teaserfigure}\includegraphics[width=\textwidth]{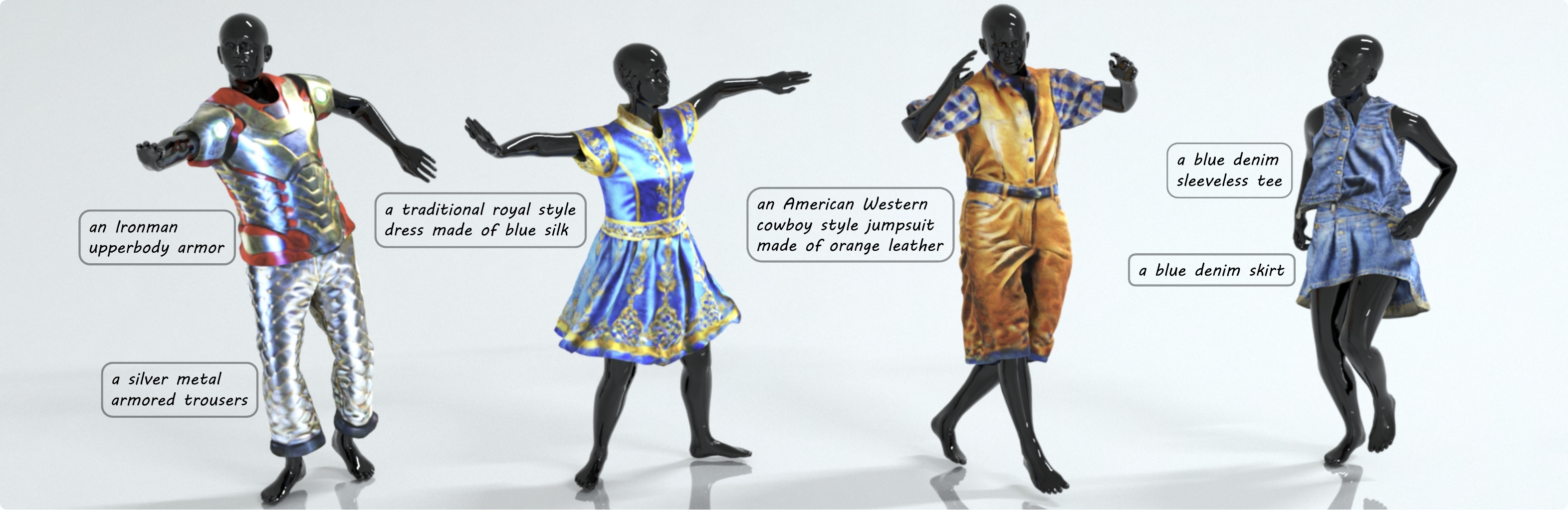}
\caption{
    GarmentDreamer is a garment synthesis framework for customizing simulation-ready high-quality textured garment meshes from text prompts.
}
  \label{fig:teaser}
\end{teaserfigure}

\maketitle

\section{Introduction}   \label{sec:intro}

The creation of 3D digital garments is crucial in graphics and vision, driven by their extensive applications in fashion design, virtual try-on, gaming, animation, virtual reality, and robotics. Nevertheless, the conventional pipeline for 3D garment creation which encompasses sketching and modeling, followed by UV mapping, texturing, shading and simulation using commercial software \cite{MAYA, CLO3D, Style3D} demands substantial manual effort. This process results in significant time and labor costs.

With the advancement of diffusion-based generative models \cite{long2023wonder3d, poole2022dreamfusion}, 3D garment generation from text and images has flourished. Two primary methods have emerged. The first method, as explored in prior works \cite{he2024dresscode, liu2023towards}, starts by reconstructing 2D sewing patterns and subsequently generating 3D garments from these patterns. The second method involves generative models that directly predict the distribution of 3D target shapes based on image and text inputs, without relying on 2D sewing patterns \cite{sarafianos2024garment3dgen, srivastava2024wordrobe, yu2023surf, tang2023dreamgaussian}. However, the former approach necessitates a vast amount of paired training data between sewing patterns and corresponding text or images \cite{liu2023towards}. The latter approach, while simpler, encounters issues such as multi-view inconsistency \cite{sarafianos2024garment3dgen} and a lack of high-fidelity details, which often require additional post-processing for downstream simulation tasks \cite{li2023subspace, yu2022meshtaichi, lan2024mil2, li2023diffavatar}. Thus, generating simulation-ready, textured garments with high-fidelity details remains challenging.

Recognizing the advantages and limitations of both traditional pipeline and modern generative models, our goal is to create high-fidelity, simulation-ready textured garments with appropriately placed openings for the head, arms, and legs. We aim to achieve details comparable to that of the traditional pipeline. In this paper, we focus on \emph{full-piece garment generation} without relying on 2D sewing patterns, which is sufficient for many graphics applications.

To achieve our goals, we leverage image/text-conditioned diffusion models. Several challenges revealed by prior methods must be addressed:
(1) Many avatar generators \cite{liao2023tada, huang2023humannorm, xu2023seeavatar} directly produce fused cloth-human models with watertight meshes. They require separation from humans and complex modifications to introduce openings for the head, arms, and legs for downstream tasks.
(2) While some high-quality non-watertight garments are generated by deforming template meshes guided by multi-view images \cite{sarafianos2024garment3dgen}, predicting unsigned distance fields (UDF) via diffusion models \cite{yu2023surf}, or optimizing meshes through differentiable simulators \cite{li2023diffavatar}, these approaches often lack detailed and realistic geometrical features or complex textures. This limitation stems from the inherent challenges in 3D shape diffusion.
(3) Deforming garment geometry solely based on multi-view images predicted by diffusion models can lead to inconsistency \cite{chen2024v3d}. Additionally, refining textures in UV space can result in over-saturated, blocky artifacts \cite{tang2023dreamgaussian, chen2023text2tex}.

To address these challenges, we introduce GarmentDreamer, a 3DGS \cite{kerbl20233d} guided garment synthesis method for simulation-ready, non-watertight garments featuring diverse geometry and intricate textures. We first leverage diffusion models and physical simulations to obtain a smooth garment template and generate corresponding Gaussian kernels via \ac{sds} loss \cite{tang2023dreamgaussian}. Subsequently, we exploit the estimated normal map and RGBA information from 3DGS as guidance in our garment augmentation module to deform meshes using a coarse-to-fine optimization approach. In the coarse stage, we refine garment contour together with neck, arm, waist, and leg openings, and then in the fine stage, multi-scale details are created under the proposed guidance. Compared with guidance from generated multi-view images, multi-view consistent guidance extracted by Gaussian kernels creates more high-quality geometry and texture details. An implicit Neural Texture Field (NeTF) is then reconstructed and subsequently augmented by \ac{vsd} loss to offer high-quality garment textures. Compared with baking Gaussian kernels into a UV map directly, our texture extraction strategy offers more consistent results. Our contributions include:
\vspace{-3px}
\begin{itemize}[leftmargin=*]
    \item We introduce a novel 3D garment synthesis method using diffusion models with 3DGS as a reference to generate simulation-ready garments from text prompts. 
    \item We propose a new garment deformer module using normal-based and RGBA-based guidance in course-to-fine garment mesh refinement stages to generate diverse, high-quality garments with complex geometrical details.
    \item We show implicit Neural Texture Fields (NeTF) can be reconstructed and be fine tuned by \ac{vsd} loss to generate high-quality garment textures.
    \item We conduct comprehensive qualitative and quantitative experiments to evaluate the superior performance of GarmentDreamer as compared to prior methods.
\end{itemize}

\section{Related Work}  \label{sec:related}

Our work focuses on generating realistic and diverse wearable simulation-ready 3D garment meshes from text prompts.
\begin{figure*} [t]
 \includegraphics[width=1.0\textwidth]{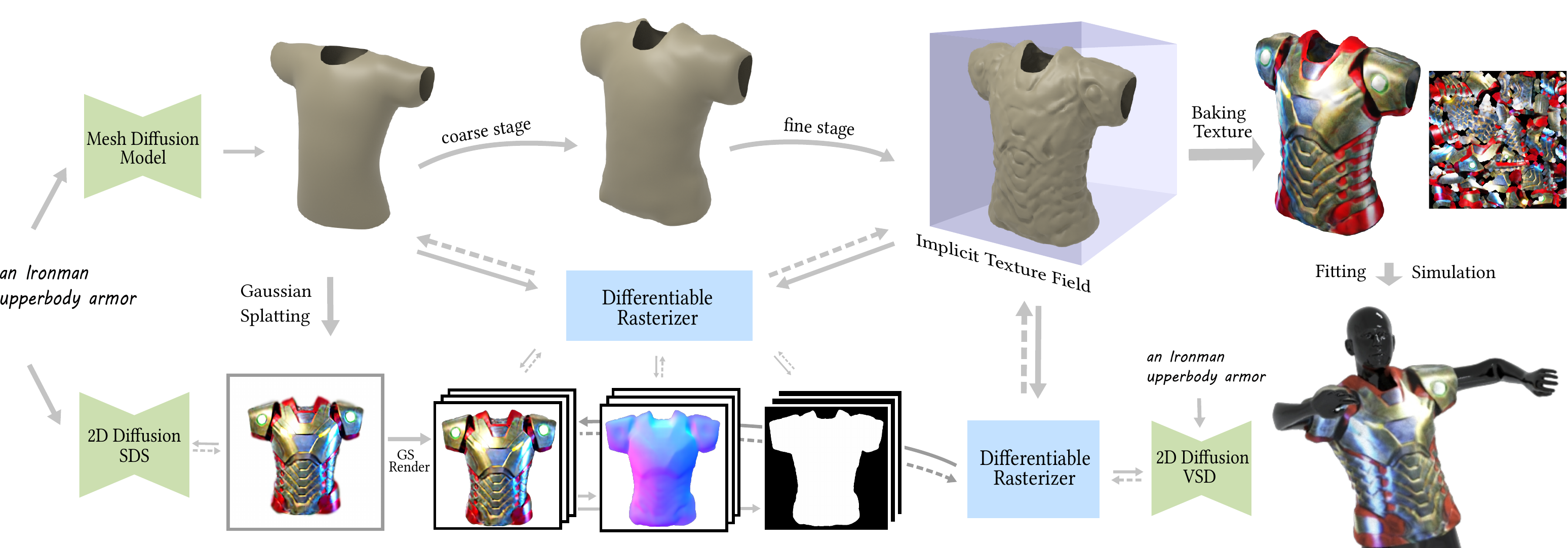}
  \centering
    \caption{Starting with text prompts, we generate a garment template mesh using a diffusion model. We then optimize a 3DGS from the template and text. Using RGBA, normal maps, and masks, we guide a two-stage deformation to refine the mesh into the final shape. Finally, we reconstruct and optimize an implicit texture field via \ac{vsd}, producing high-quality textured garment meshes that can be applied to downstream simulation/animation tasks. \vspace{-10px}}
  \label{fig:pipeline}
\end{figure*}

\paragraph{Diffusion-based 3D Generation}
For 3D generation, many work distill 2D pre-trained diffusion models  via \ac{sds} loss \cite{poole2022dreamfusion, zhou2024headstudio} and \acf{vsd} \cite{wang2024prolificdreamer, zhangtowards}, or exploit 3D diffusion models to directly generate 3D representations such as point cloud \cite{luo2021diffusion}, \acf{nerf} \cite{hong2023lrm, shue20233d}, mesh \cite{qian2023magic123, long2023wonder3d}, SDF \cite{shim2023diffusion, chou2023diffusion}, \acf{udf} \cite{yu2023surf}, DMTets \cite{shen2021deep}, and 3D\ac{gs} \cite{yi2023gaussiandreamer, tang2023dreamgaussian}. To capture rich surface details and high-fidelity geometry of generated 3D shapes, normal maps \cite{long2023wonder3d, huang2023humannorm, liu2023sherpa3d, li2023sweetdreamer}, depth maps \cite{qiu2023richdreamer}, pose priors \cite{zhang2024avatarverse}, skinned shape priors \cite{kim2023chupa} have been adopted as guidance modules for 3D digital avatar \cite{xu2023seeavatar, jiang2023mvhuman, li2023diffavatar}, garment \cite{he2024dresscode, sarafianos2024garment3dgen}, and scene synthesis tasks \cite{tang2023diffuscene, vuong2024language, lee2024semcity}. 

\paragraph{3D Garment Synthesis}
Traditional 3D garment creation usually begins with 2D sewing patterns in commercial fashion design software, necessitating significant labor and time costs \cite{sarafianos2024garment3dgen}. To automate 3D garment generation, learning-based methods have been employed to infer garment shapes from text prompts, images, and videos \cite{he2024dresscode, liu2023towards, yu2023surf, casado2022pergamo, chen2023gm, xiang2023drivable, zhang2023getavatar, de2023drapenet, li2023diffavatar, srivastava2024wordrobe, sarafianos2024garment3dgen, he2023sketch2cloth, shen2020gan}. 
However, many methods focused on clothed human synthesis \cite{xiang2023drivable, avatarfusion, wang2023disentangled, zhang2023getavatar, chen2023gm} typically generate garments fused together with digital human models, which restricts them to basic skinning-based animations and requires nontrivial work to separate the garments from the human body. 
In contrast, our work focuses on separately wearable geometry. Other closely related work include  \cite{li2023diffavatar} which also generates high-quality simulation-ready clothes at the expense of creating clothing templates by artists and precise point clouds by scanners. 

Some recent methods \cite{liu2023towards, he2024dresscode} generate non-watertight garments with sewing patterns, while ours generates a one-piece cloth similarly to \citet{sarafianos2024garment3dgen}. Although one-piece 3D clothing is not suitable for manufacturing, it is still sufficient for many graphics downstream tasks. Note that \cite{sarafianos2024garment3dgen} develops methods to improve multi-view consistency in image generation similarly to \cite{long2023wonder3d}, while our method mitigates this challenge through a novel 3DGS guidance.

\paragraph{Garment Refinement} 
Garment refinement involves  diversification \cite{kim2023chupa}, draping \cite{de2023drapenet}, wrinkle generation \cite{muller2010wrinkle, chen2023deep}, stylization \cite{sarafianos2024garment3dgen} and other techniques to enhance diversity and realism. Traditional methods optimize energy or geometric constraints to directly edit meshes \cite{sorkine2004laplacian, botsch2007linear}. Recent learning-based approaches modify latent geometry and textures, potentially with conditioned diffusion \cite{kim2023chupa, srivastava2024wordrobe, de2023drapenet, he2024dresscode}, and decode latent features to generate refined meshes. To modify garment meshes from text prompts, Textdeformer \cite{gao2023textdeformer} represents mesh deformation through Jacobians and exploits global CLIP features as guidance. Garment3DGen \cite{sarafianos2024garment3dgen} deforms garment geometry using 2D image guidance. However, these RGB-based or CLIP-based methods have a tendency to prefer modifying 2D textures over 3D geometric structures. On the other hand, Surf-D \cite{yu2023surf} edits garment geometry with sketch conditions but lacks texture synthesis. \citet{kim2023chupa} utilize normal maps as references to carve clothed humans. \citet{worchel2022multi} exploit neural deferred shading results to modify mesh surfaces. Inspired by these works \cite{kim2023chupa, worchel2022multi}, our approach explores both normal maps and RGBA image features with neural deferred shading to guide mesh deformation, creating realistic garment geometry and textures that can be directly applied to downstream simulation and animation tasks.

\vspace{-5px}
\section{Method}  \label{sec:method}
\autoref{fig:pipeline} overviews our method. Given text $\mathcal{T}$, GarmentDreamer generates wearable textured garment meshes $\mathcal{F}_{g}$ with neck, arm, waist, and leg openings. Starting with $\mathcal{T}$, we first generate garment template $\mathcal{F}_{t}$ based on predicted \ac{udf}s in \autoref{sec:template}. Then we optimize a 3D\ac{gs} representation in \autoref{sec:3dgs} based on $\mathcal{T}$ and $\mathcal{F}_{t}$. Leveraging 3D\ac{gs} guidance, we design a two-stage training in \autoref{sec:refine} to refine  $\mathcal{F}_{t}$ into the final garment shape $\mathcal{F}_{g}$. Finally, we generate high-quality textures by optimizing an implicit Neural Texture Field (NeTF) ${\Phi}$ augmented with variational \ac{sds} loss in \autoref{sec:texture}.

\vspace{-5px}
\subsection{Garment Template Mesh Generation}  \label{sec:template}

Human clothing generation can utilize strong inductive biases, as garments within the same category tend to have similar topology and overall orientation. Thus, it is reasonable to warm-start our optimization using a template and modify it. The starting point is to generate a template mesh $\mathcal{F}_{t}$ from text $\mathcal{T}$. We use the meshes from Cloth3D  \cite{bertiche2020cloth3d} and SewFactory \cite{liu2023towards} as our dataset to train a mesh diffusion model.  Leveraging the efficiency and effectiveness of generation in latent space \cite{rombach2022high}, we represent the geometry information of the garment mesh in a compact vector form by extracting its latent code. 
\vspace{-5px}
\paragraph{Simulation-based Preprocessing}
A straightforward approach to obtaining the garment latent space is to train an autoencoder with high-quality garment meshes \cite{yu2023surf}. However, through experiments, we found that the autoencoder struggles to capture high-frequency geometric details, such as overly dense wrinkles. Additionally, some meshes in the dataset exhibit self-intersections due to these noisy details, resulting in unsatisfactory reconstruction. We believe these high-frequency details are not only challenging for the autoencoder to learn but also redundant for the purpose of template generation. Recognizing that we only need a warm-start template and will obtain geometric details in later phases, we propose a simulation-based data preprocessing step that involves smoothing the garment meshes using a physics-based cloth simulator. Specifically, we set the rest bending angle between every two adjacent triangles to zero. By minimizing the bending energy in quasi-static Finite Element Method (FEM) simulation steps \cite{li2021codimensional} together with a Neo-Hookean stretching energy, each garment mesh transforms to a rest state with minimal high-frequency wrinkles, without altering the overall characteristic shape.

\vspace{-5px}
\paragraph{Garment Latent Space}
To encode the 3D garment template meshes $\mathcal{F}_{t}$ into latent space, we utilize the Dynamic Graph CNN (DGCNN) \cite{wang2019dynamic}, producing a 64-dimensional garment latent code $\gamma$. To reconstruct the garment geometry from the latent code, we employ a Multilayer Perceptron (MLP) with Conditional Batch Normalization \cite{de2017modulating} as the decoder. This decoder processes the latent code alongside a set of query points, which are sampled from the input meshes and their surroundings during training, and randomly sampled in a canonicalized space during testing. The decoder then predicts the \ac{udf} values for these points. We use both distance loss and gradient loss as suggested by \citet{de2023drapenet} in training the autoencoder.
\vspace{-5px}
\paragraph{Latent Diffusion}
Building upon the garment latent space, we train a latent diffusion model to predict the latent code $\gamma$ conditioned on the garment category specified in the text prompt $\mathcal{T}$, such as a skirt or T-shirt. Using the garment decoder alongside a set of query points, we obtain the garment UDF field. Finally, the desired garment template mesh $\mathcal{F}_{t}$ is extracted from the predicted UDF using MeshUDF \cite{guillard2022meshudf}.
\vspace{-3px}

\subsection{3D Gaussian Generation}  \label{sec:3dgs}

Given a generated template mesh that is smooth and relatively generic, our next step is to generate proper guidance models to enhance it with details. While prior work has utilized multi-view image generators, they suffer from risks of inconsistency \cite{chen2024v3d}. We turn to 3D radiance field representations to completely mitigate potential multi-view inconsistency risks.

To this end we adopt 3DGS \cite{kerbl20233d}, which exploits 3D anisotropic Gaussian kernels to reconstruct 3D scenes with learnable mean $\mu$, opacity $\sigma$, covariance $\Sigma$, and spherical harmonic coefficients $\mathcal{S}$. We utilize the garment template mesh $\mathcal{F}_{t}$ with the same text prompt $\mathcal{T}$ offering color, material, and pattern descriptions to generate 3D Gaussian kernels for further guiding geometry refinement and texture generation. Similar to the query points used in the garment latent code decoder, the initial 3D Gaussians are randomly sampled at the surroundings of the template mesh surface. Following \citet{yi2023gaussiandreamer}, 
we optimize Gaussian kernels using \ac{sds} loss \cite{poole2022dreamfusion} with a frozen 2D diffusion model $\phi$ conditioning on text prompts $\mathcal{T}$. The rendered image is produced by a differentiable renderer $g$ with the parameters of 3DGS $\theta$, notated as $\mathbf{x} = g(\theta)$. The formula for computing the gradient to guide the updating direction of $\theta$ is:
$
    \nabla_\theta \mathcal{L}_{\mathrm{SDS}}(\phi, \mathbf{x}=g(\theta)) \triangleq \mathbb{E}_{t, \epsilon}\left[w(t)\left(\hat{\epsilon}_\phi\left(\mathbf{z}_t ; \mathcal{T}, t\right)-\epsilon\right) \frac{\partial \mathbf{x}}{\partial \theta}\right],
$
where $\hat{\epsilon}_\phi\left(\mathbf{z}_t ; \mathcal{T}, t\right)$ is the score estimation function, predicting the sampled noise $\hat{\epsilon}$ given the noisy image $\mathbf{z}_t$, text prompt $\mathcal{T}$, and noise level $t$; $w(t)$ is a weighting function.

To render a view, the 3D Gaussian kernels are projected to 2D screen space based on their $z$-depth. Each pixel color $C_{p}$ is computed by $\alpha$-blending these 2D Gaussians from near to far: $C_{p} = \sum_{i \in N} p_{i}\alpha_{i}\prod_{j=1}^{i-1} (1 - \alpha_{j})$, where $p_{i}$ is the evaluated color by spherical harmonics viewed from the camera to the kernel's mean; $\alpha_{i}$ is the product of the kernel's opacity and 2D Gaussian weight evaluated at the pixel coordinate. We render RGB images and masks of Gaussian kernels from 24 views for the next stage. The masks are generated using a step function with an empirical threshold $\vartheta$ applied to the opacity $\sigma \in [0,1]$ of each Gaussian kernel.

\vspace{-2px}
\subsection{Garment Geometry Deformer} \label{sec:refine}

We propose a two-stage training process to optimize the geometry of a garment to match the multi-view guidance generated from 3DGS. The first coarse stage utilizes masks to optimize the contour of the garment's geometry. The second fine stage employs RGB renderings and normal maps to enrich local details. Additionally, we use the mesh generated in the coarse stage to detect hole regions and maintain garment openings for wearability.
\vspace{-2px}
\subsubsection{Coarse Stage}
Given the template mesh $\mathcal{F}_t$ and a camera view $C_i$, we use a differentiable rasterizer to generate the current mask. We use the following mask loss to guide the contour optimization:
\vspace{-2px}
\begin{equation}
    \mathcal{L}_{\text{M}} (\bm{v}) = \frac{1}{|\mathcal{I}|}\sum_{i \in \mathcal{I}}\operatorname{MSE}(R_M(\mathcal{F}_t, C_i), M_i),
    \vspace{-2px}
\end{equation}
where the optimization variable $\bm{v}$ is the concatenation of all vertex positions, $\mathcal{I}$ is the set of camera views, $R_M$ is the differentiable contour rasterizer empowered by Nvdiffrast \cite{Laine2020diffrast}, and $\{M_i\}$ is the ground truth mask guidance.

Optimization under merely $\mathcal{L}_{\text{M}}$ is stochastic and unstable. One reason is that the vertices inside the masks can move freely without changing the output mask. Considering that the template mesh is smooth, we can maintain the smooth surface and stabilize the deformation process during the coarse stage by imposing constraints on the surface curvature. This is achieved through a normal-consistency loss $\mathcal{L}_{\text{NC}}$ and a Laplacian loss $\mathcal{L}_{\text{L}}$ as in \citet{worchel2022multi}:
\begin{equation}
\small
  \mathcal{L}_{\text{NC}}(\bm{v}) = \frac{1}{N}\sum_{j \sim k} (1 - \bm{n}_j \cdot \bm{n}_k)^2, \ \ 
 \mathcal{L}_{\text{L}}(\bm{v}) = \frac{1}{M} \sum_{j \sim k}w_{jk}\|\bm{v}_j - \bm{v}_k\|^2,
\end{equation}
where $\bm{n}_i, \bm{n}_j$ are adjacent face normals, $N$ is the number of adjacent face pairs, $\bm{v}_i, \bm{v}_j$ are adjacent vertex positions, $M$ is the number of adjacent vertex pairs, and $\{w_{jk}\}$ are the Laplacian edge weights.

In summary, the loss function for the coarse stage is the weighted sum of these losses. During our implementation, we found that without any additional constraints to be applied other than the aforementioned losses, hole regions can be preserved during the deformation process and follow the deformation of the mesh surface. Furthermore, the boundaries of the hole regions coincide with the boundaries of the garment rendered by Gaussian rendering. This provides us with good intermediate results. By utilizing the deformed hole regions, we can continue to preserve the openings and the garment's wearability in the next deformation stage.

\vspace{-2px}
\subsubsection{Fine Stage}
While contour mask guidance can optimize the garment mesh to match the overall shape of the generated 3DGS, it does not encourage the generation of local geometric details characterized by local displacement variations, which are invisible to masks. On the other hand, RGB renderings do not provide sufficient geometric information since shading effects can be baked into textures without changing the geometry. This is also a limitation of 3DGS, where the geometric distribution of Gaussian kernels may not align with the actual objects. 

However, the RGB renderings of 3DGS provide rich visual information, which encapsulates a wealth of geometric details, allowing for a possible understanding of the underlying geometry structure. To utilize the RGB information to enrich geometry details, we use the neural shader module from \citet{worchel2022multi}, which is an implicit shading field $S(\bm{x}, \bm{n}, \bm{d})$ that maps a query position along with its normal and view direction to an RGB colour. The module is combined with the same differentiable rasterizer $R$ to render RGB images. Given a camera view $C_i$ with the camera center $\bm{c}_i$, we jointly optimize the shader's parameters $\theta$ and garment vertices $\bm{v}$ using the following RGB loss:
\begin{equation}
  \mathcal{L}_{\text{RGB}} (\bm v, \theta) = \frac{1}{|\mathcal{I}|}\sum_{i \in \mathcal{I}}\operatorname{L_1}(S(\tilde{\bm v}_{jk}, \tilde{\bm n}_{jk}, \tilde{\bm v}_{jk} - \bm c_i), (I_i)_{jk}),
\end{equation}
where $\{I_i\}$ is the ground truth RGB images rendered from the generated 3DGS, $\tilde{\bm{n}} = R(\{\bm{n}_j\}, \mathcal{F}_t, C_i)$ and $\tilde{\bm{v}} = R(\{\bm{v}_j\}, \mathcal{F}_t, C_i)$ are rasterized vertex normals and positions, respectively. 

Simultaneously, we observed that the neural shader sometimes brings noise and artifacts, like protrusions and indentations, into geometry. To address this issue, we need to maintain the necessary geometry details like wrinkles while removing the noise. We propose using normal estimation models to obtain estimated normal maps used for additional guidance, which can capture wrinkles and the overall normal information of the garment. The ground truth normal maps $\{N_i\}$ are inferred from the rendered RGB images from the generated 3DGS by a pre-trained normal estimator, Metric3D \cite{yin2023metric3d}. We then use the following normal loss to guide the geometry optimization:
\begin{equation}
    \mathcal{L}_{\text{N}} (\bm{v}) = \frac{1}{|\mathcal{I}|}\sum_{i \in \mathcal{I}}\operatorname{L_1}(R(\{\bm{n}_j\}, \mathcal{F}_t,  C_i), N_i),
\end{equation}
where $\{\bm{n}_j\}$ is the set of vertex normals. 

Furthermore, the normals at the garment hole regions are not reliable since the renderings from the 3DGS at these regions are blurry and directly applying the normal map at the hole regions to the mesh deform process can result in the closure of the original openings. We propose using the coarse-stage mesh as the hole guidance to maintain the openings. we detect the hole region by checking the dot product between the garment surface normal and the camera direction. Specifically, the hole region mask within a rendered image is defined as
\begin{equation}
  (\tilde{M}^H_i)_{jk} = \tilde{\bm{n}}_{jk} \cdot (\tilde{\bm{v}}_{jk} - \bm{c}_i) > 0.
\end{equation}
We use the following hole loss to maintain the holes initially present in the template mesh after the coarse stage:

\begin{equation}
\mathcal{L}_{H}(\bm v) = \frac{1}{|\mathcal{I}|}\sum_{i\in I} \operatorname{MSE}(\tilde{M}^H_i, M^H_i),
\end{equation}
where $M^H_i$ is the hole mask at the beginning of the fine stage. However, the mask values are boolean, which are not differentiable. We manually skip the gradient of the binarization by letting
\begin{equation}
  \frac{\partial \mathcal{L}^H}{\partial \{\tilde{\bm n}_{jk} \cdot (\tilde{\bm v}_{jk} - c_i)\}} = \frac{\partial \mathcal{L}^H}{\partial (\tilde{M}^H_i)_{i, j}},
\end{equation}
which can provide correct gradient directions.

In summary, the loss function for the fine stage is the weighted sum of $\mathcal{L}_{\text{M}}, \mathcal{L}_{\text{NC}}, \mathcal{L}_{\text{L}}, \mathcal{L}_{\text{N}}, \mathcal{L}_{\text{H}}, \mathcal{L}_{\text{RGB}}$. These losses will align the geometry with the 3DGS appearance but also maintain the openings of the garment to ensure it is wearable and simulation-ready.

\begin{figure*} [t]
 \includegraphics[width=1.0\textwidth]{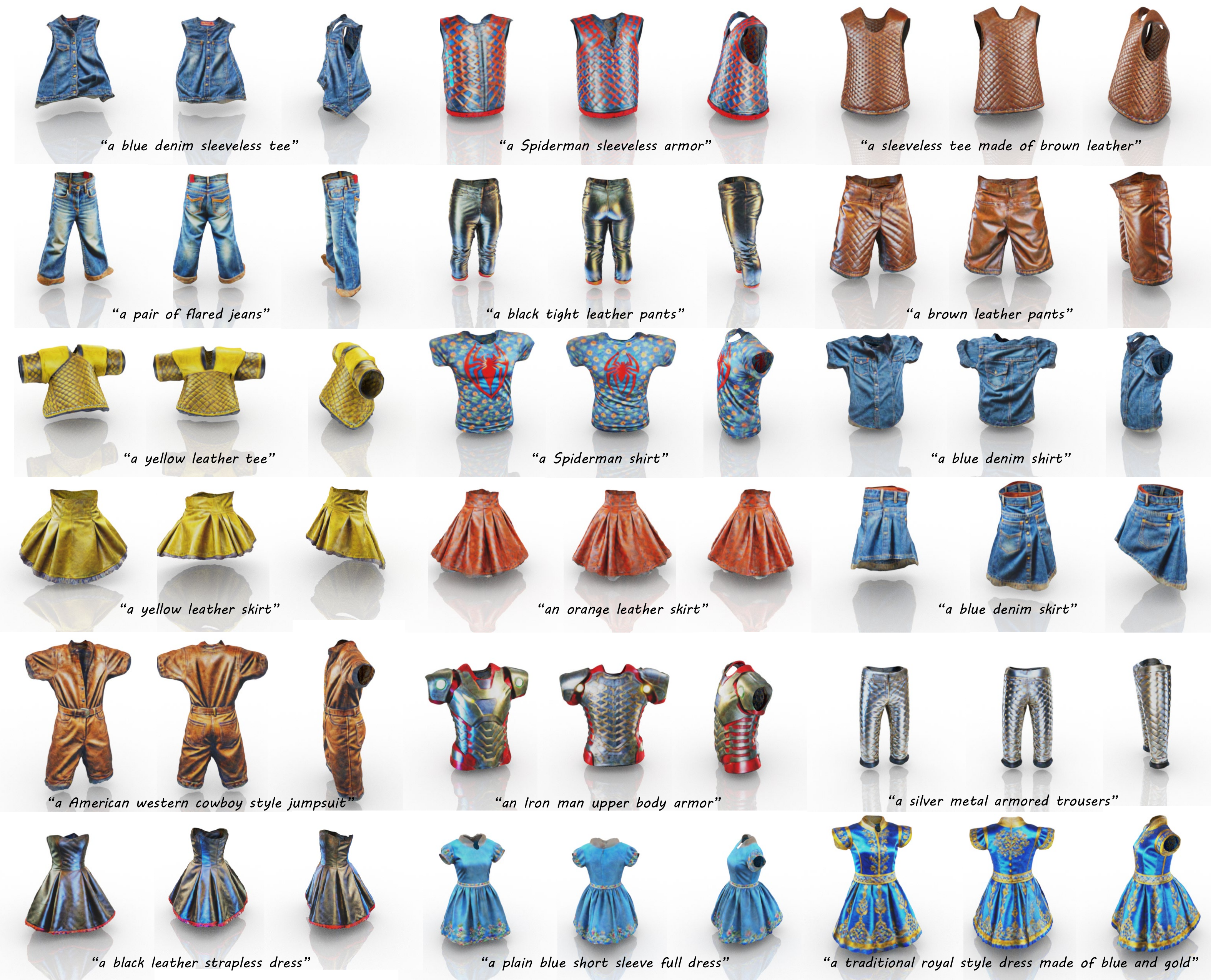}
  \centering
  \vspace{-0.6cm}
  \caption{\textbf{Garment Gallery}. We showcase a gallery of textured garment meshes of different clothing categories generated by GarmentDreamer. We refer to the supplementary material for closer observations of these garments and more generation results.} 
  \label{fig:showcases}
\end{figure*}

\subsection{Texture Synthesis} \label{sec:texture}

The final step of GarmentDreamer is to generate detailed textures for garment meshes. Since our goal is to synthesize simulation-ready garments with high-fidelity, high-quality textures, simply using the neural shader previously mentioned to extract vertex colors cannot achieve this with a limited number of mesh vertices. Notably, the RGB guidance from the 3DGS retain the original high-quality texture information in the form of multiple images. Following Instant-NGP \cite{muller2022instant}, we propose using an implicit Neural Texture Field (NeTF) $\Phi(\bm{x})$ to reconstruct these multiview images into high-quality textures, which can then be easily extracted onto a UV map. The functionality is equivalent to the aforementioned neural shader without dependency on surface normals and view directions, which maps a query position $\bm{x}$ to its color. We optimize NeTF using the following loss function:

\vspace{-10px}
\begin{equation}
    \mathcal{L}_{\operatorname{T}}(\omega) = \frac{1}{|\mathcal{I}|}\sum_{i \in \mathcal{I}}\operatorname{L_1}(\Phi(\tilde{\bm v}_{jk}), (I_i)_{jk}),
    \vspace{-5px}
\end{equation}
where $\omega$ is the parameters of the NeTF $\Phi$ and $\tilde{\bm v}$ is the rasterized template mesh vertex positions. Note that this texture reconstruction step does not require a large number of mesh vertices, allowing us to achieve both high-quality UV textures and maintain the mesh's simulation-ready property without any extra post-process.

For better visual quality and enhanced texture details, we utilize \acf{vsd} \cite{wang2024prolificdreamer} to fine-tune the implicit texture fields. This process involves forwarding the NeTF to project a 2D image from a random view and feeding this projected image into the \ac{vsd} framework to compute a perceptual loss. By backpropagating this loss, we optimize the parameters of NeTF implicitly and improve the overall texture quality.

After reconstruction and fine-tuning, we can easily query the color of any mesh point $\bm p$, which is baked onto a texture map.
\vspace{-5px}

\subsection{Garment Simulation}
We use the SMPL-X model \cite{pavlakos2019expressive}  as the articulated human base to simulate garment dynamics. Initially, we manually adjust the garment on the human template by scaling, translating, and altering the human pose. This rough fitting often results in numerous penetrations. To address this, we employ Position-Based Dynamics (PBD) \cite{muller2007position} to push the garment mesh outside the human body. During this process, the garments are equipped with high stretching and bending stiffness to preserve the initial shape. Subsequently, we apply Codimensional Incremental Potential Contact (CIPC) \cite{li2021codimensional} to simulate the garments on a human motion sequence, using manually selected physical parameters. The human mesh is treated as a moving boundary condition. We use CIPC because it can ensure penetration-free results, which is essential for accurate garment simulations. Note that the state generated by PBD is merely an initial feasible state for CIPC. The rest pose of the garment remains as originally generated, though adjustments in the scale of the pose can be made to modify the tightness of the fit.

\vspace{-2px}
\section{Experiments}
In this section, we conduct a thorough evaluation of GarmentDreamer across various garment categories, providing both quantitative and qualitative comparisons with other state-of-the-art 3D generation methods. We also present ablation studies to highlight the effectiveness of the key components in our pipeline.

\vspace{-5px}
\subsection{Showcases}

\begin{figure*}
  \centering
 \includegraphics[width=\textwidth]{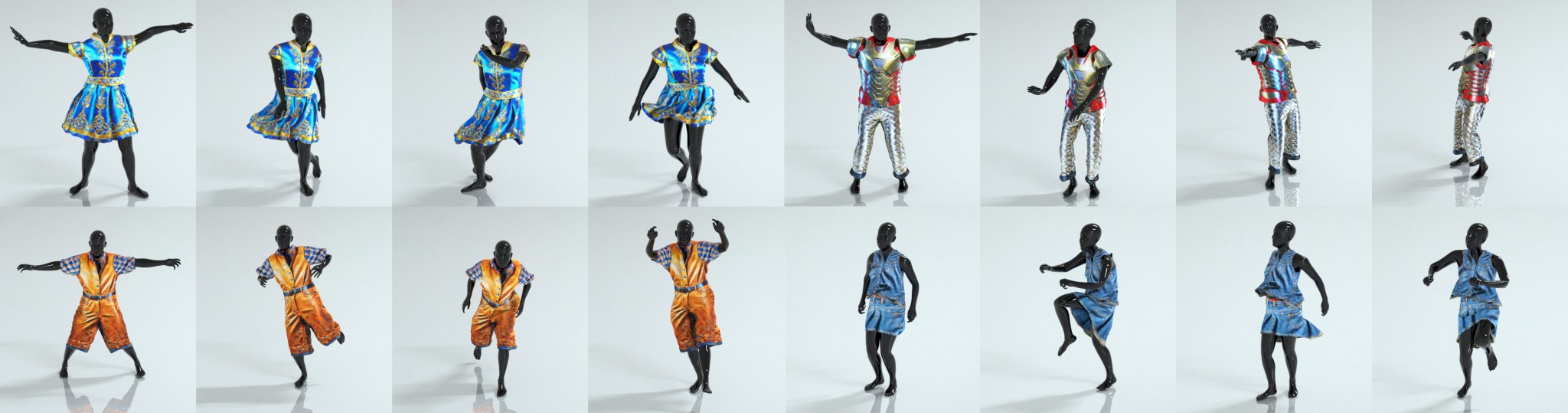}
 \vspace{-0.6cm}
 \caption{We use the SMPL-X mesh sequence to drive the dynamics of our generated garments. The utilization of CIPC resolves frictional collisions and self-collisions effectively and guarantees non-penetrative results in garment simulations.}
 \label{fig:animation}
\end{figure*}

We visualize a range of 3D textured garment meshes generated by GarmentDreamer, encompassing diverse categories such as dresses, trousers, shirts, skirts, tees, and jumpsuits. GarmentDreamer not only produces intricate geometric details that align with pattern descriptions but also generates a variety of high-quality textures based on material descriptions. As shown in \autoref{fig:showcases}, GarmentDreamer successfully creates skirts and tees with consistent denim textures. Additionally, the generated non-watertight wearable garments are ready for direct application in simulations and animations. We showcase four dancing sequences in \autoref{fig:teaser}.  Human motion are generated by Mixamo\footnote{\url{https://www.mixamo.com/}}. The animation sequences are in \autoref{fig:animation}.

\begin{figure*} [t]
 \includegraphics[width=1.0\textwidth]{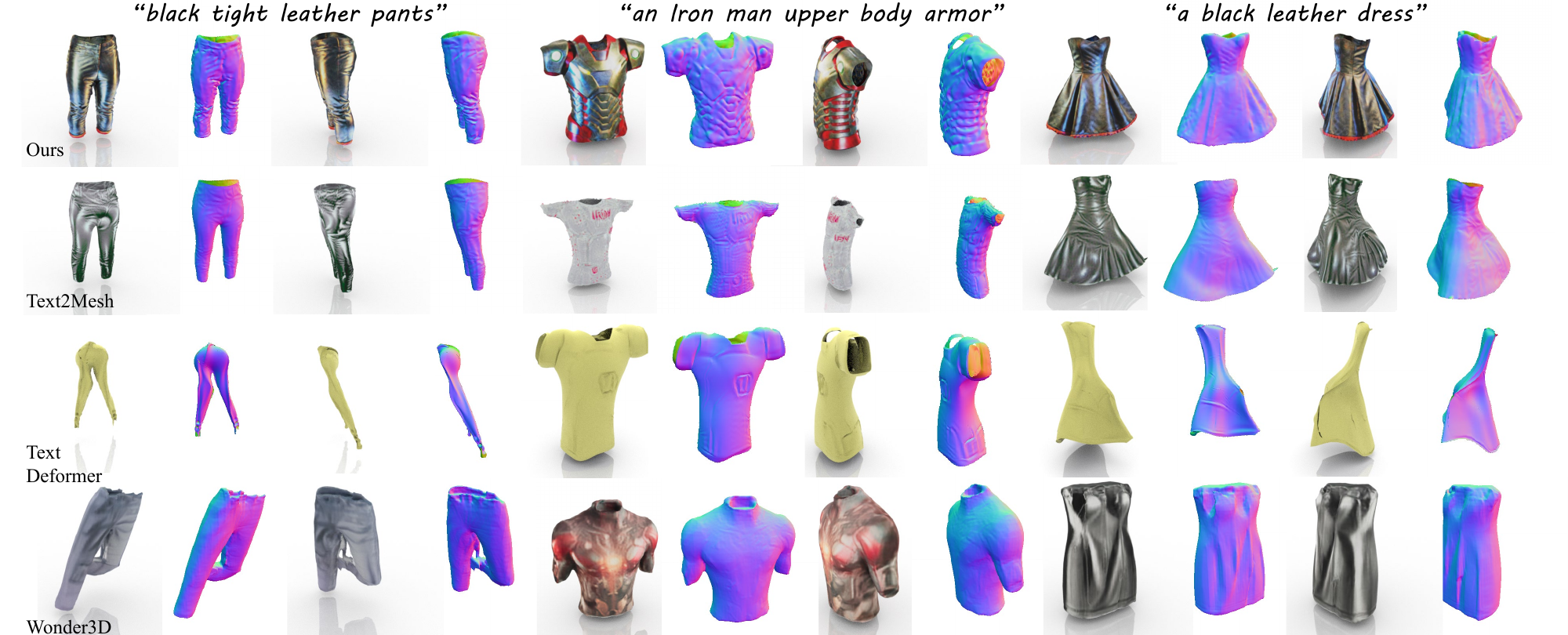}
  \centering
  \caption{\textbf{Normal Comparisons}. We visualize the normal maps for a better comparison of garment geometry between GarmentDreamer and other methods. Our proposed method generates visually plausible garment meshes, featuring finer geometric details such as natural wrinkles and smooth boundaries. \vspace{-10px}
  }
  \label{fig:normal_comparison}
\end{figure*}

\vspace{-2px}
\subsection{Comparison} \label{sec:com}
We compare GarmentDreamer against several state-of-the-art 3D generation methods: Text2Mesh \cite{michel2022text2mesh}, TextDeformer \cite{gao2023textdeformer}, and Wonder3D \cite{long2023wonder3d}. Text2Mesh and TextDeformer aim to optimize and deform an initial mesh to the desired shape indicated in the text prompt. We use the same template mesh for GarmentDreamer and these deformation-based methods to ensure fairness. To compare our method with Wonder3D, which reconstructs 3D meshes from single-view images, we use DALLE-3 \cite{betker2023improving} to generate the images from text prompts as its input. To comprehensively compare GarmentDreamer with these three methods, we generate 21 distinct types of garments with each method, including shirts, dresses, skirts, and pants.

\vspace{-8px}
\subsubsection{Quantitative Comparison}
In the absence of a standardized metric for 3D generation quality, we focus on measuring the consistency between the generated garments and their text prompt inputs. We render each garment from 36 different views and compute the average CLIP similarity score between these rendered images and the corresponding text prompts. Given that TextDeformer does not handle textures, we render its results using a default color. Typically, text-to-2D/3D works utilize the vanilla CLIP model \cite{radford2021learning} for evaluating text-to-image alignment. However, this model is optimized for general subjects and lacks specificity for garment evaluation. To address this, we employ FashionCLIP \cite{chia2022contrastive}, a model similar to CLIP but fine-tuned on a fashion dataset, making it more appropriate for our purposes.

We report the comparison results in \autoref{tab:quantitative comparison}, including both the FashionCLIP Similarity Score (FCSS) and the CLIP Similarity Score (CSS). Our method, which is specifically designed for 3D garments, outperforms all baselines in terms of text-garment alignment. Furthermore, our approach demonstrates faster performance compared to the two deformation-based baselines. Unlike joint optimization between CLIP-based guidance and mesh deformation, our method benefits from a decoupled and efficient 3DGS step followed by the mesh optimization step.

\begin{table}[t]
\centering
\caption{\textbf{Quantitative Comparisons.} Our approach outperforms deformation-based and generative methods on both FashionCLIP similarity score (FCSS) which is pretrained on fashion datasets and vanilla OpenAI CLIP similarly score (CSS), supports generating openings on clothing as well as textures, and runs faster than prior deformation-based methods. \vspace{-5px}} 
\label{tab:quantitative comparison}
\small{
\begin{tabular}{p{0.55in}cccccc}
\hline
\textbf{Methods} &\textbf{\textbf{FCSS}}& \textbf{CSS}&\textbf{Wearable}& \textbf{Texture}& \textbf{Runtime}&\\ \hline
Text2Mesh  & 0.3396& \cellcolor{Goldenrod}{0.2655}& $\checkmark$ & $ \checkmark$ & $\sim$ 20 mins&\\ 
TextDeformer & 0.2367& 0.1657& $\checkmark$ & & $\sim$ 35 mins &\\
Wonder3D & \cellcolor{Goldenrod}{0.3402} & 0.2509&  &$\checkmark$ &  \cellcolor{Apricot}{$\sim$ 4 mins} &\\
\hline
 Ours& \cellcolor{Apricot}{0.3413}& \cellcolor{Apricot}{0.2731}&  $\checkmark$ &  $\checkmark$& \cellcolor{Goldenrod}{$\sim$ 15mins}&\\
\hline
\end{tabular}
\vspace{-10px}
}
\end{table}

\vspace{-4px}
\subsubsection{Qualitative Evaluation}
\begin{figure*} [t]
 \includegraphics[width=1.0\textwidth]{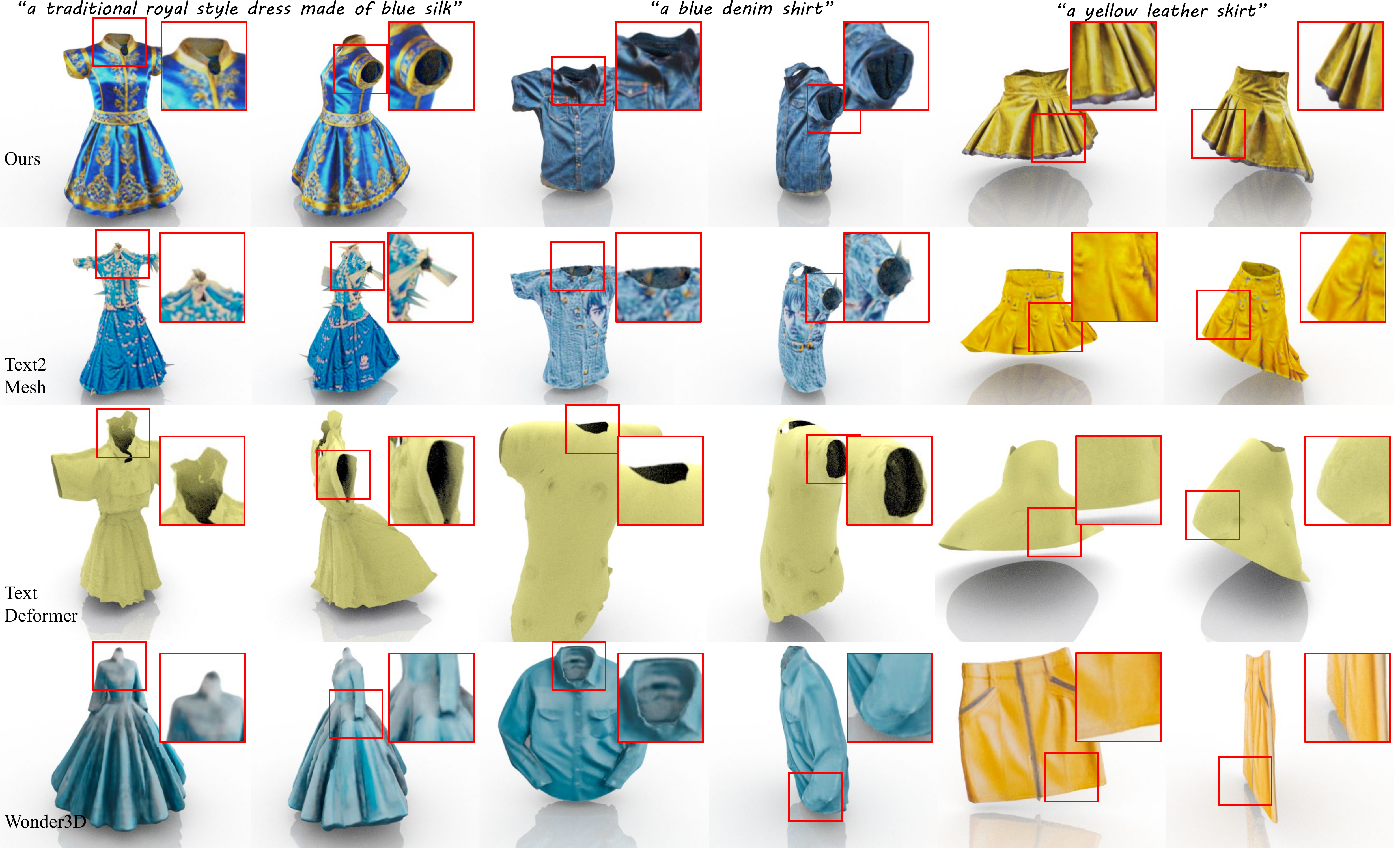}
  \centering
  \vspace{-0.8cm}
  \caption{\textbf{Qualitative Comparisons}. While baseline methods either produce unrealistic geometric artifacts, e.g. spikes and excessive smoothness, or non-garment textures, GarmentDreamer excels in generating high-quality, simulation-ready non-watertight garments with detailed textures and fine wrinkle details.   \vspace{-10px}}
\label{fig:zoomin_comparison}
\end{figure*}

GarmentDreamer ensures that the garment meshes maintain their geometric integrity and exhibit rich, detailed textures, making them suitable for high-quality visual applications. We visualize results of GarmentDreamer and other baselines in \autoref{fig:zoomin_comparison} and \autoref{fig:normal_comparison}. The meshes produced by Text2Mesh often appear distorted with spiky artifacts due to the direct optimization of all vertex coordinates. TextDeformer alleviates this issue by parameterizing deformation as Jacobians, but it tends to miss high-frequency details, resulting in overly smooth geometry. Wonder3D relies heavily on input images and frequently generates garments with closed sleeves or necklines due to limited garment-specific knowledge. In contrast, our method produces non-watertight, simulation-ready garments with realistic textures, enabling seamless downstream tasks like animation and virtual try-on. 

\begin{figure*} [t]
 \includegraphics[width=1.0 \textwidth]{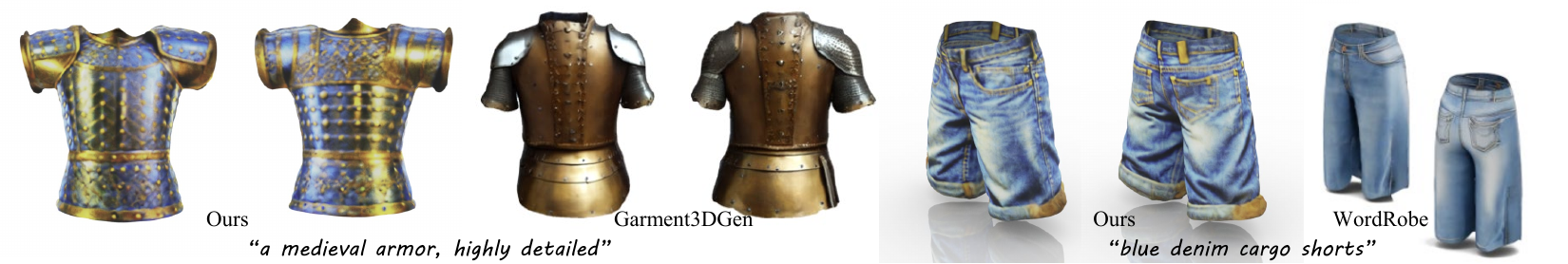}
  \centering
  \vspace{-0.6cm}
  \caption{We compare with images from Garment3DGen \cite{sarafianos2024garment3dgen} and WordRobe \cite{srivastava2024wordrobe}. Ours show high-quality details.} 
  \label{fig:qua}
\end{figure*}

We further compare with Garment3DGen \cite{sarafianos2024garment3dgen} and WordRobe \cite{srivastava2024wordrobe} using the same text prompts; see \autoref{fig:qua}. Cargo shorts generated by \cite{srivastava2024wordrobe} have fewer details in grain lines, hemline, seam allowance, and side seams than ours. Armor created by Garment3DGen \cite{sarafianos2024garment3dgen} aligns textures with image guidance, while ours generate detailed high-quality geometric structures.

\vspace{-6px}
\subsection{Ablation} 
\begin{figure*} [t]
 \includegraphics[width=1.0 \textwidth]{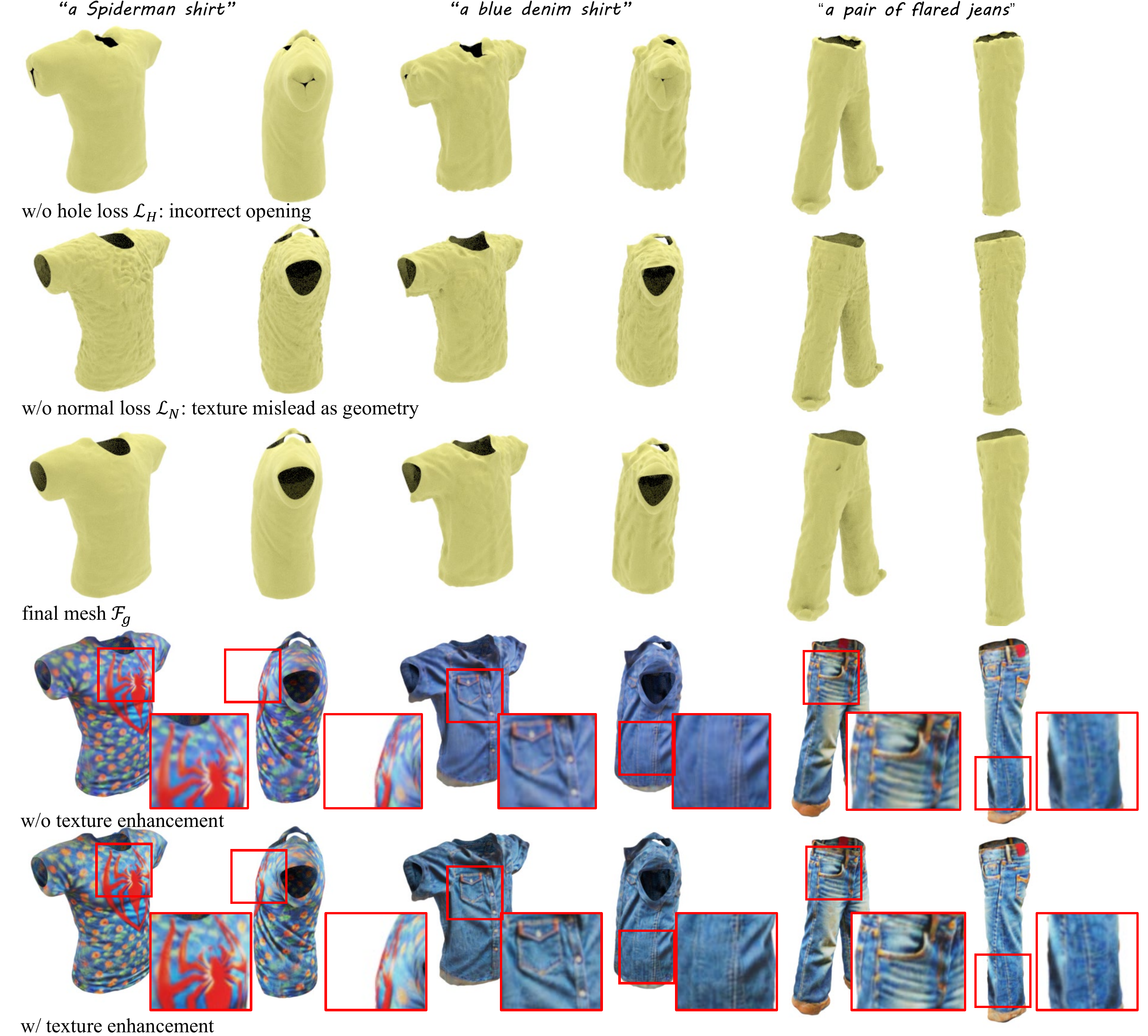}
  \centering
  \vspace{-0.6cm}
  \caption{\textbf{Ablation Study}. From top to bottom are results of GarmentDreamer without hole loss $\mathcal{L}_H$, normal loss $\mathcal{L}_N$, final mesh without texture, texture without refinement, and final textured mesh respectively. Hole loss $\mathcal{L}_H$, normal loss $\mathcal{L}_N$ offers clear openings, noise-free surface of the generated garment mesh. Texture enhancement provides high-quality details in hemline, seam allowance, grain line and wrinkles. } 
  \label{fig:ablation}
\end{figure*}

In \autoref{fig:ablation}, we conduct ablation for key components in GarmentDreamer, using the same 21 generated garments in \autoref{sec:com}.

\vspace{-0.2cm}
\paragraph{Hole Loss}
We first examine the effectiveness of the proposed hole loss $\mathcal{L}_H$, a crucial component of our proposed 3DGS-guided mesh deformer, which ensures the openings stay non-closed during optimization. Without this, mesh deformation tends to enclose the arm and head holes, resulting in unwearable garment meshes.

\vspace{-0.2cm}
\paragraph{Normal Loss}
Normal loss $\mathcal{L}_N$ ensures noise-free meshes. For the spiderman shirt example in \autoref{fig:ablation}, the garment surface without normal loss guidance tends to be messy and noisy, and incorrectly represents Spiderman texture as geometry structures, which is against fabric design in the real world. 

\vspace{-0.2cm}
\paragraph{NeTF Enhancement} Our proposed NeTF enhancement via \ac{vsd} facilitates better capture of fabrication attributes. 
As shown in \autoref{fig:ablation}, it could be observed that without NeTF enhancement, the pocket of the shirt and the waistband are blurry. In contrast, texture enhancement leads to high-quality fabrication details, such as fold lines, grain lines, the seam allowance of the shirt, and clear fly piece, hemline, seam allowance, and side seam of the jumpsuit, which are crucial in traditional garment design.

\vspace{-5px}
\section{Limitations and Future Work}

There are several avenues for future research. Firstly, our method takes minutes rather than seconds for each generation process. Improving its efficiency and scalability is essential for its application to large-scale garment collections. Secondly, integrating differentiable physics-based dynamic simulations during the generation process may provide additional important physical guidance to improve the realism of the generated garments. Furthermore, parameterizing the geometry with 2D sewing patterns could offer multiple benefits: they facilitate a seamless connection to the manufacturing process, ensure a closer match between real and simulated clothing, and provide a more intuitive design workflow for traditional fashion designers. Lastly, similar to other SDS-based 3D generation approaches, our method currently bakes some lighting effects such as specular highlights and shadows into the color map. Separating these effects via learning Physically-Based Rendering (PBR) materials would likely  improve the quality and realism of the results.

\bibliographystyle{ACM-Reference-Format}
\bibliography{ref}

\end{document}